\documentclass[letterpaper,10pt,conference]{ieeeconf}

\IEEEoverridecommandlockouts
\usepackage{url}
\usepackage[dvipsnames]{xcolor}
\usepackage{wrapfig}
\usepackage{booktabs}

\usepackage{amsmath}
\usepackage{float}
\usepackage{optidef}
\usepackage{setspace}
\usepackage{graphicx}
\usepackage{colortbl}
\usepackage{mathrsfs}
\usepackage{amssymb}
\usepackage{accents}

\usepackage{nicefrac}
\usepackage{algorithm}
\usepackage{dirtree}
\usepackage{cases}
\usepackage[noend]{algpseudocode}

\usepackage{flushend}
\usepackage{multicol}

\usepackage{pifont}
\newcommand{\cmark}{\textcolor{ForestGreen}{\ding{51}}}%
\newcommand{\xmark}{\textcolor{red}{\ding{55}}}%

\makeatletter
\newcommand\fs@spaceruled{\def\@fs@cfont{\bfseries}\let\@fs@capt\floatc@ruled
  \def\@fs@pre{\vspace{0.4\baselineskip}\hrule height.8pt depth0pt \kern2pt}%
  \def\@fs@post{\vspace{-0.4\baselineskip}\kern2pt\hrule\relax\vspace{-12pt}}%
  \def\@fs@mid{\kern2pt\hrule\kern2pt}%
  \let\@fs@iftopcapt\iftrue}
\makeatother

\usepackage{lipsum}
\usepackage{pgfplots}
\pgfplotsset{tick label style={font=\bfseries\color{white!15!black}},}
\pgfplotsset{compat=newest}
\newlength\figH
\newlength\figW

\usepackage{cite}

\usepackage{hyperref}
\hypersetup{
    colorlinks=true,
    urlcolor=magenta
}

\definecolor{melon}{rgb}{0.99, 0.74, 0.71}
\definecolor{my_gray}{HTML}{616B85}

\usepackage{listings}

\definecolor{dkgreen}{rgb}{0,0.6,0}
\definecolor{gray}{rgb}{0.5,0.5,0.5}
\definecolor{mauve}{rgb}{0.58,0,0.82}

\usetikzlibrary{plotmarks}

\lstdefinelanguage{Julia}%
  {morekeywords={abstract,break,case,catch,const,continue,do,else,elseif,%
      end,export,false,for,function,immutable,import,importall,if,in,%
      macro,module,otherwise,quote,return,switch,true,try,type,typealias,%
      using,while},%
   sensitive=true,%
   alsoother={\$},%
   morecomment=[l]\#,%
   morecomment=[n]{\#=}{=\#},%
   morestring=[s]{"}{"},%
   morestring=[m]{'}{'},%
}[keywords,comments,strings]%

\newif\ifanon

\definecolor{limogreen}{RGB}{0,158,115}
\DeclareRobustCommand{\method}{%
    \textbf{\textcolor{limogreen}{RAYA}}\hspace{0.2em}%
}

\definecolor{darkgreen}{rgb}{0.0,0.45,0.0}
\definecolor{refblue}{RGB}{0,90,160}

\title{\LARGE \bfseries
\method: Learning Where and When to Intervene for Robot Recovery}

\ifanon
    \author{Anonymous Authors}
\else
    \author{
        Ishaan Mahajan$^{1,2}$,
        Charles Chen$^{1,2}$,
        Frederike D\"umbgen$^{1}$, and
        Brian Plancher$^{2}$%
        \thanks{This project was supported by the National Science Foundation (Awards 2411369, 2535096). Any opinions, findings, conclusions, or recommendations expressed in this material are those of the authors and do not necessarily reflect those of the funding organizations.}%
        \thanks{$^{1}$College of Engineering, Carnegie Mellon University, Pittsburgh, PA, USA.}%
        \thanks{$^{2}$Dartmouth College, Hanover, NH, USA.}%
        }
\fi

\begin{document}

\maketitle
\thispagestyle{empty}
\pagestyle{empty}

\begin{abstract}
A robot can predict failure and still be unable to prevent it. By the time a safety mechanism reacts, the nominal plan may already have spent the control authority that recovery requires, and fixed task priorities may block whatever response remains.
Our key insight is that both aspects are decided inside the controller. Recoverability must inform actions while they are chosen rather than veto them afterward, and task objectives must be adapted as recoverability shrinks.
Building on this, we present \method, a hybrid learned--analytic framework that places a learned finite-horizon recoverability margin inside an optimal controller with hard constraints and pairs it with a bounded learned scheduler that shifts task weights to facilitate recovery.
Across 7,200 simulation episodes per controller spanning quadrotor and autonomous-vehicle benchmarks, \method not only improves survival rates, but also transfers the learned components zero-shot to unseen trajectories, disturbances, plant shifts, and friction layouts.
We developed an embedded realization of \method and deployed it onboard a 35g Crazyflie quadrotor. Across 40 combined hardware flights under wind with either aerodynamic mismatch or an unmodeled 40\% motor-command loss, each of three baselines fails in all trials, while \method completes 10/10 six-cycle missions. Project Website: \url{https://raya-control.github.io/}.

\end{abstract}

\begin{figure*}[t]
    \centering
    \includegraphics[width=0.77\textwidth]{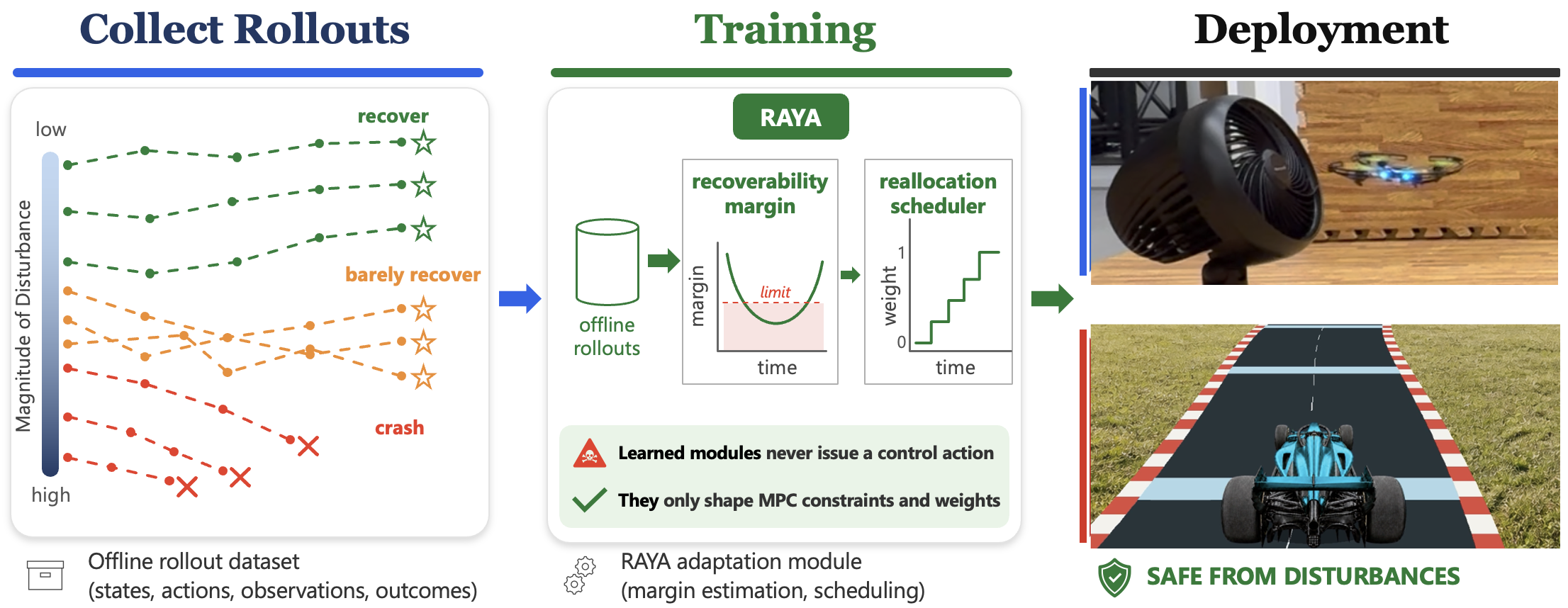}
    \vspace{-3pt}
    \caption{
    \textbf{\method~learns when recoverability is deteriorating and how the task
    should respond.}
    Closed-loop rollouts train a finite-horizon recoverability predictor and a
    bounded reallocation scheduler. At deployment, their outputs jointly modify a predictive control problem where the recoverability margin shapes the constraints,
    while the scheduler reweights selected tracking objectives before the control
    action is optimized.
}
    \vspace{-8pt}
    \label{fig:limo_framework}
\end{figure*}

\section{Introduction}
\label{sec:intro}
Unexpected aerodynamic forces or a shifted payload may consume a quadrotor's thrust and attitude authority~\cite{hanover2021performance}.
Unmapped surfaces may similarly lack sufficient friction for a ground vehicle's control authority~\cite{vaskov2024friction}. 
Such failures are always preceded by a point past which no admissible input can enable recovery.
Thus, knowing that a system is unsafe is not the same as being able to save it. 
It is important to consider \emph{whether the controller can identify that recoverability is deteriorating while sufficient time and actuation authority remain}. 

Common runtime safety architectures are often layered over the outputs of a nominal planning and control stack. Unfortunately, this means that when the safety layer changes the action, the plant may execute an input that leads to unplanned areas of the state space. Under tight input limits, such a safety mechanism may even be triggered too late to enable recovery. A rich family of safety filters implement this design, including control barrier function (CBF) quadratic programs~\cite{ames2016control}, Hamilton--Jacobi (HJ) safety supervisors~\cite{fisac2018general,bansal2017hamilton,bansal2021deepreach,borquez2024safety}, and predictive safety filters~\cite{wabersich2021predictive,tearle2021predictive}, alongside learned constructions that extend safety values to failure modes that are hard to model analytically~\cite{robey2020learning,lindemann2021learning,ma2022learning,lindemann2024learning,so2024train,knoedler2025safety,dacs2026robust,lin2025one,dawson2023safe}.

Safety information can instead enter predictive optimization while an action is selected. Discrete-time CBF constraints and layered CBF--MPC architectures impose safety conditions inside the prediction horizon~\cite{zeng2021safety,grandia2021multi}. Recent methods even embed learned neural barriers in sampling-based model predictive control (MPC)~\cite{yin2025safe,kaypak2026control}, constrain MPC with learned reachability-based safety values~\cite{wang2026cooptimizing}, or adapt predictive models online~\cite{zhou2025simultaneous}. These approaches make risk visible, but they do not enable the planner and controller to reason about which task objectives can or should be modified in response. Thus, \emph{recoverability often competes with task objectives during deployment}.

A third line of work addresses the task sacrifice directly. Context-dependent MPC weights~\cite{zarrouki2024safe}, jointly adapted objective and barrier parameters~\cite{sabouni2024reinforcement}, priority-based constraint relaxation~\cite{prignoli2025priority}, and recoverability-oriented barrier constructions~\cite{wijayatunga2026learning} allow priorities to move at runtime, while reset policies and dedicated recovery controllers learn when to retreat, switch, or return to nominal operation~\cite{eysenbach2017leave,thananjeyan2020safety,thananjeyan2021recovery,vats2025recoverychaining}. These methods establish the value of adaptive priorities and of learning from failure experience, yet each resolves the question in one of two incomplete ways: either priorities adapt without a recoverability signal to drive them, or recovery is achieved through a discrete handoff to a separate action-generating policy. To our knowledge, no prior method jointly addresses \textit{where} recoverability should enter the optimizer and \textit{when} selected task objectives should yield, all within one planning and control framework.

We introduce \method{} (Recoverability-Aware Yielding of Authority), a hybrid learned--analytic control framework that uses a learned recoverability signal (Fig.~\ref{fig:limo_framework}). A neural model is trained on narrow platform-specific datasets and evaluated zero-shot on unseen trajectories, disturbance processes, plant shifts, and friction layouts. This model predicts the minimum task-specific safety margin over a finite closed-loop recovery horizon from signals available to the deployed controller, and a local approximation of that margin reshapes predictive feasibility while selecting the current action. The same margin also enters a lightweight bounded scheduler, together with deployable state and actuator-risk features, so objectives that can yield are temporarily downweighted while recovery-critical channels retain or gain weight. The prediction dynamics, state and input bounds, and platform-specific constraints remain unchanged. Thus, learning tracks when recovery is being lost and how strongly selected task objectives should yield, while the underlying optimal controller decides the final action subject to hard constraints. 

Across 7,200 simulation episodes per controller spanning quadrotor and autonomous-vehicle (AV) benchmarks, \method not only improves survival rates, but also transfers the learned components zero-shot to unseen trajectories, disturbances, plant shifts, and friction layouts. We developed an embedded realization of \method and deployed it onboard a 35g Crazyflie quadrotor. Across 40 combined hardware flights under wind with either aerodynamic mismatch or an unmodeled 40\% motor-command loss, each of three baselines fails in all trials, while \method completes 10/10 six-cycle missions. We release our project open-source. 

Overall, this work makes three main contributions.
\begin{itemize}
    \item \textbf{Hybrid Learned--Analytic Recovery.}
    A learned finite-horizon margin changes predictive feasibility, a bounded scheduler conditionally reweights selected task objectives, and the underlying optimal controller remains the action-generator subject to hard constraints.

    \item \textbf{Out-of-Distribution (OOD) Transfer.}
    Without retraining, the learned components generalize to unseen tasks, disturbance distributions, plant shifts, and friction layouts, and the same controller design transfers across aerial and ground embodiments.

    \item \textbf{Recovery without a Compute Bottleneck.}
    An embedded realization retains \method{}'s performance while running on an MCU on a 35g Crazyflie quadrotor.
\end{itemize}

\section{Background}
\label{sec:background}

At control time \(t\), the controller receives a state estimate~\(\hat{x}_t\) of the state \(x_t\in\mathcal{X}\subseteq\mathbb{R}^{n_x}\).
 Given reference states \(x^{\mathrm{ref}}_{k|t}\) and inputs
\(u^{\mathrm{ref}}_{k|t}\), MPC solves 
\begin{subequations}
\label{eq:nominal_mpc}
\begin{align}
\min_{\mathbf{x}_t,\mathbf{u}_t}\quad&
\sum_{k=0}^{N-1}
\left(
\|x_{k|t}-x^{\mathrm{ref}}_{k|t}\|_{Q_0}^{2}
+
\|u_{k|t}-u^{\mathrm{ref}}_{k|t}\|_{R_0}^{2}
\right)
\nonumber\\[-1mm]
&+
\|x_{N|t}-x^{\mathrm{ref}}_{N|t}\|_{Q_f}^{2}
\label{eq:nominal_cost}
\\
\mathrm{s.t.}\quad&
x_{0|t}=\hat{x}_t,
\\
&
x_{k+1|t}
=
A_{k|t}x_{k|t}
+
B_{k|t}u_{k|t}
+
c_{k|t}, 
\\
&
x_{k|t}\in\mathcal{X},
\qquad
u_{k|t}\in\mathcal{U}, 
\qquad k\in[0, N-1].
\end{align}
\end{subequations}
Here \(A_{k|t}\), \(B_{k|t}\), and \(c_{k|t}\) define the local affine
prediction model at stage \(k\), \(N\) is the finite horizon,
\(\|a\|_Q^2:=a^\top Q a\), and \(Q_0\), \(R_0\), and \(Q_f\) are objective
weights. The set \(\mathcal{U}\subseteq\mathbb{R}^{n_u}\) denotes the
admissible input set. We use
\(\mathbf{x}_t=\{x_{0|t},\ldots,x_{N|t}\}\) and
\(\mathbf{u}_t=\{u_{0|t},\ldots,u_{N-1|t}\}\).
Only the first optimized input \(u_{0|t}^{\star}\) is applied before
replanning.

For implementation, the problem can be expressed in the
operator-splitting form
\begin{subequations}
\label{eq:split_mpc}
\begin{align}
\min_{\xi,z}\quad&
\frac{1}{2}\xi^\top H_t\xi
+
g_t^\top\xi
+
\Psi_t(z),
\\
\mathrm{s.t.}\quad&
E_t\xi=e_t,
\\
&
\xi-z=0.
\end{align}
\end{subequations}
Here \(\xi\) stacks the decision variables \(\mathbf{x}_t\) and \(\mathbf{u}_t\), \(z\) is the auxiliary
variable used for separable projections, and \(E_t\xi=e_t\) collects the
initial-state and linearized dynamics constraints. The matrix \(H_t\)
contains the quadratic objective weights induced by \(Q_0\), \(Q_f\), and
\(R_0\), while \(g_t\) contains the corresponding reference-dependent linear
terms. The function \(\Psi_t\) collects the separable projections
associated with the state and input constraints~\cite{boyd2011distributed}.

We solve \eqref{eq:split_mpc} using the ADMM-based, Riccati-structured
implementation of TinyMPC
\cite{tinympc, conic-tinympc}.
For this work, the useful consequence of formulation \eqref{eq:split_mpc} is that it
separates two interfaces through which recovery information can modify the
controller during action selection. Additional recoverability conditions enter
through \(\Psi_t\), while reweighting selected state-tracking objectives
changes the corresponding objective terms in \(H_t\) and \(g_t\). The prediction
dynamics and pre-existing hard state and input constraints remain unchanged.

We distinguish post hoc from in-solver intervention. 
A post hoc safety filter modifies \(u_{0|t}^{\star}\) only after the nominal
problem \eqref{eq:nominal_mpc} has been solved. The predicted trajectory is
therefore generated without accounting for that intervention, and a requested
correction may be infeasible once the nominal plan has approached the input
limits.
An in-solver intervention instead modifies the optimization problem used to select
\(u_{0|t}^{\star}\), allowing the predicted trajectory and current
action to adapt jointly to
task objectives and recoverability conditions. \method follows this architecture through both
interfaces: recoverability modifies predictive feasibility,
while objective reallocation changes which task objectives are
prioritized. Section~\ref{sec:method} develops these two interventions.

\section{\method}
\label{sec:method}
Figure~\ref{fig:limo_framework} summarizes the offline training and an online control step of \method{}\!. Offline, narrow, platform-specific, closed-loop rollouts provide worst-future-margin labels for training the learned recoverability predictor. Then, held-out rollouts calibrate its offset, and optimize a bounded objective reallocation scheduler in closed loop.
Online, the learned predictor estimates the recoverability margin from the current state and recent history. Its local linearization enters the MPC problem of Sec.~\ref{sec:background} as an additional constraint in \(\Psi_t\). 
The same margin then feeds a bounded objective reallocation scheduler, \(\bar w_t\in[0,1]\), that reweights task objective channels, with the corresponding quadratic and reference-dependent objective terms, \(H_t\), \(g_t\), recomputed accordingly. 
The modified optimal control problem is solved once and \(u_{0|t}^{\star}\) is applied zero-shot to unseen scenarios. 
Neither learned component generates a control action, and the robot's dynamics, input bounds, and hard constraints remain unchanged.

\subsection{Learning When Recovery Is Being Lost}
\label{sec:recoverability}

Let \(m:\mathcal{X}\rightarrow\mathbb{R}\) denote a signed instantaneous
margin, with larger values indicating greater separation from failure. Given
an offline closed-loop rollout, we label each time step by the minimum future
margin over a recovery horizon \(H_{\mathrm{rec}}\),
\begin{equation}
    y_t=
    \operatorname{smin}_{0\leq j\leq H_{\mathrm{rec}}}m_{t+j},
    \label{eq:recoverability_target}
\end{equation}
where  \(m_t=m(x_t)\), \(j\) is the future-step offset, and \(\operatorname{smin}\) denotes either the hard
minimum or its smooth log-sum-exp approximation. We learn a regressor
\(h_{\psi}\) that predicts this finite-horizon margin from deployable features
\(\zeta_t\),
\begin{equation}
    y_t \approx h_{\psi}(\zeta_t):=f_{\psi}(\zeta_t)-T.
    \label{eq:learned_margin}
\end{equation}
Here \(f_{\psi}\) is a neural network and \(T\) is calibrated on held-out closed-loop rollouts against false-safe predictions. A \(\pm50\%\) sweep of
\(T\) changes at most one of 4,200 quadrotor outcomes and spans 1.5 points of
AV survival. The features \(\zeta_t\) use only deployable signals: altitude,
vertical velocity, tilt, attitude margin, angular rate, and actuation authority
for the quadrotor; tracking error, vehicle response, slip and model residuals,
saturation, curvature, and short histories for the AV. Neither receives the
realized disturbance or a preview.

\subsection{Putting Recoverability Inside the Solver}
\label{sec:insolver}

The learned margin modifies the optimization problem with which \(u_{0|t}^{\star}\) is selected, rather than correcting the action post hoc. 
Let \(\bar x_{k|t}\) denote the linearization point at prediction
stage \(k\), and let \(\zeta_{k|t}(x_{k|t})\) denote the corresponding feature vector, with fixed observed history quantities. We first form a first-order approximation of
\(x\mapsto h_{\psi}(\zeta_{k|t}(x))\) about \(\bar x_{k|t}\),
\begin{equation}
\scalebox{0.92}{$
\widehat h_{k|t}(x)=\;
h_{\psi}\!\left(\zeta_{k|t}(\bar x_{k|t})\right)
+\nabla_x h_{\psi}\!\left(\zeta_{k|t}(\bar x_{k|t})\right)^{\!\top}
(x-\bar x_{k|t}).
$}
\label{eq:barrier_linearization}
\end{equation}

We define \(\tau_{b,k|t}\) as the recoverability threshold at prediction stage \(k\).  The threshold \(\tau_{b,k|t}\) is separate from the calibration offset
\(T\). The offset \(T\) calibrates the learned recoverability prediction,
while \(\tau_{b,k|t}\) sets the margin enforced by MPC. Its platform-specific parameterization uses only quantities available at deployment, is fixed before evaluation, and is shared across the matched controller ablations. 

We can then impose the affine recoverability constraint as:
\begin{equation}
    \widehat h_{k|t}(x_{k|t})
    \geq
    \tau_{b,k|t}.
    \label{eq:limo_halfspace_hard}
\end{equation}

When preserving feasibility of the underlying MPC problem is desired, we use a softened form with nonnegative slack \(s_{k|t}\): 
\begin{equation}
    \widehat h_{k|t}(x_{k|t}) + s_{k|t}
    \geq
    \tau_{b,k|t},
    \qquad
    s_{k|t}\geq0, 
    \label{eq:limo_halfspace_soft}
\end{equation}
and we add the penalty \(\lambda_{k|t}s_{k|t}^{2}\), with \(\lambda_{k|t}\geq0\), to the nominal objective \eqref{eq:nominal_cost}.

The learned constraint therefore augments rather than replaces the pre-existing constraints already present in MPC. For example, for the quadrotor, we retain an altitude--descent guard and a minimum-collective constraint alongside the softened recoverability constraint~\eqref{eq:limo_halfspace_soft}, while for the AV, the learned recoverability constraint is combined with the existing trajectory-tube constraint.

Although \(h_{\psi}\) predicts a finite-horizon quantity, the corresponding
constraint need not be imposed at every MPC stage. Let \(\mathcal{K}_t\)
denote the set of prediction stages at which the learned recoverability
constraint is imposed. By default, we use only the first predicted stage,
\(\mathcal{K}_t=\{1\}\), in the selected controller. Sec.~\ref{sec:ablations}
compares the impact of this choice with deeper placements.

\subsection{Learning When Task Priorities Should Yield}
\label{sec:authority}

The recoverability constraint excludes or penalizes predicted states with low recoverability, but it does not specify when competing tracking objectives should yield as recoverability deteriorates. We therefore allow the reallocation scheduler to modify selected blocks of the nominal state-cost matrix \(Q_0\). 
Let \(\mathcal{E}\) denote the yieldable task channels and \(\mathcal{R}\) denote recovery-critical channels, those whose regulation becomes more important for recovery. We define
\begin{align}
Q_{\mathcal{E}}(\bar w_t)&=
\max(\eta_{\min},1-\beta_T\bar w_t)Q_{\mathcal{E}}^0,
\label{eq:task_scaling}
\\
Q_{\mathcal{R}}(\bar w_t)&=
(1+\beta_A\bar w_t)Q_{\mathcal{R}}^0.
\label{eq:recovery_scaling}
\end{align}

Here \(0<\eta_{\min}\leq1\) is the minimum fraction of nominal weight retained
on channels in \(\mathcal{E}\), while \(\beta_T\geq0\) and
\(\beta_A\geq0\) determine the amount of attenuation and amplification,
respectively. All remaining entries of \(Q_0\) are unchanged. 

The choice of \(\mathcal{E}\) and \(\mathcal{R}\) depends on the control
task. For the quadrotor, lateral position and velocity may yield while
attitude and angular-rate regulation receive greater weight. 
For the AV,
position and speed tracking may yield, while heading and steering
regulation are preserved.

The scheduler is a smoothed logistic policy where
\(\phi_t \in \mathbb{R}^{11}\) contains a bias, a reactive margin ramp,
the calibrated recoverability margin relative to its intervention threshold,
deployable state and actuator-risk features drawn from \(\zeta_t\), and the
previous scheduler output,
\begin{equation}
  \scalebox{0.93}{$
      w_t=\sigma(\theta^\top\phi_t),\quad
      \bar w_t=\alpha\bar w_{t-1}+(1-\alpha)w_t,
      \quad \alpha=0.70
  $}
\label{eq:authority_scheduler}
\end{equation}

We train the scheduler in closed loop
using episodic cross-entropy search \cite{rubinstein2004cross}. Candidate schedules are
ranked by survival, then tracking error, then mean reallocation. The learned models and scheduler
parameters are held fixed during all evaluations. The training
distributions used for the experiments are
described in Sec.~\ref{sec:evaluation_protocol}.

\subsection{Bounded-Intervention Guarantee}
Having defined both learned components, we next state what their
modifications can and cannot change in the underlying optimization.
Although the learned recoverability margin is empirical, the way \method
modifies MPC is structurally bounded. Let \(\mathcal{C}_t\) denote the
feasible set defined by the prediction dynamics and all pre-existing hard state,
input, and platform constraints at time \(t\). Since
\(\bar w_t\in[0,1]\), \(0<\eta_{\min}\leq1\), and
\(\beta_T,\beta_A\geq0\), Eqs.~\eqref{eq:task_scaling}--\eqref{eq:recovery_scaling}
imply
\begin{equation}
\scalebox{0.9}{$
\eta_{\min}Q_{\mathcal E}^0
\preceq Q_{\mathcal E}(\bar w_t)
\preceq Q_{\mathcal E}^0,
\;
Q_{\mathcal R}^0
\preceq Q_{\mathcal R}(\bar w_t)
\preceq (1+\beta_A)Q_{\mathcal R}^0 .
$}
\end{equation}
Thus, with the positive-semidefinite state and terminal costs,
\(R_0\succ0\), and \(\lambda_{k|t}\geq0\), the \method objective remains
convex for every scheduler output. Since scheduling modifies only the objective terms \(H_t\) and \(g_t\),
it leaves \(\mathcal{C}_t\) unchanged and therefore cannot relax a pre-existing
hard constraint. A hard learned recoverability constraint intersects
\(\mathcal{C}_t\) with an additional affine halfspace and therefore can only
restrict the feasible set. For the softened recoverability constraint, every
trajectory in \(\mathcal{C}_t\) remains feasible by choosing
\[
s_{k|t}
=
\bigl[\tau_{b,k|t}-\widehat h_{k|t}(x_{k|t})\bigr]_+ .
\]
Hence the softened recoverability constraint cannot make an otherwise
feasible MPC problem infeasible. 

Thus, learning changes what the optimizer prefers and what it avoids, never what the platform is permitted to do.

\section{Experiments}
\label{sec:experiments}
\begin{figure*}[t]
    \centering
    \includegraphics[width=0.90\textwidth]
        {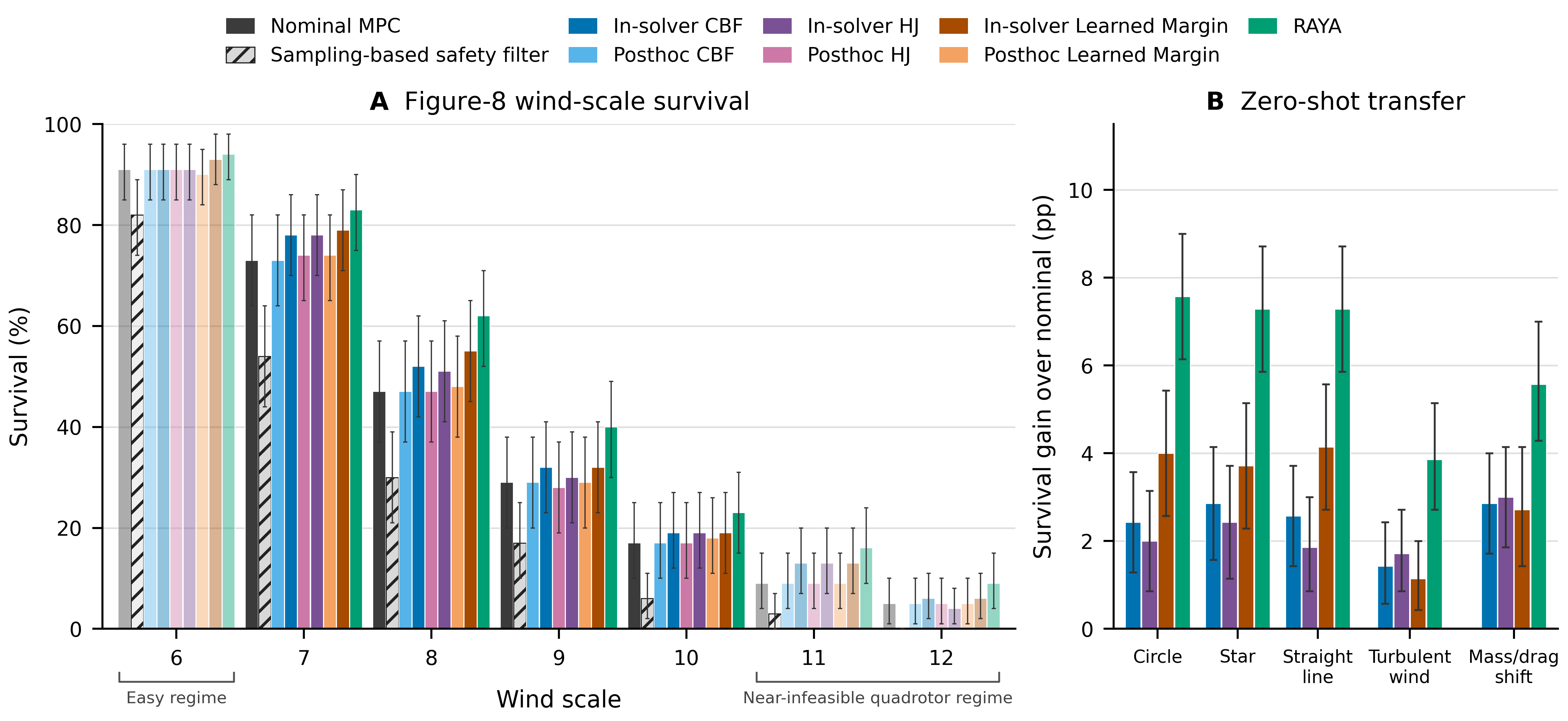}
        \vspace{-8pt}
        \caption{
            \textbf{Quadrotor disturbance robustness and zero-shot transfer.}
            \textbf{(A)} Figure-8 survival across the wind sweep for nine controllers using
            100 shared seeds per method--wind condition. Controller separation is largest
            between the benign and near-infeasible regimes.
            \textbf{(B)} Survival gain over nominal MPC for controllers that improve on unseen
            trajectories, turbulent wind, and mass/drag shift. Whiskers show 95\%
            whole-seed clustered bootstrap intervals.
        }
    \label{fig:quad_envelope}
    \vspace{-8pt}
\end{figure*}
\begin{figure*}[t]
    \centering
    \includegraphics[width=0.75\textwidth]
        {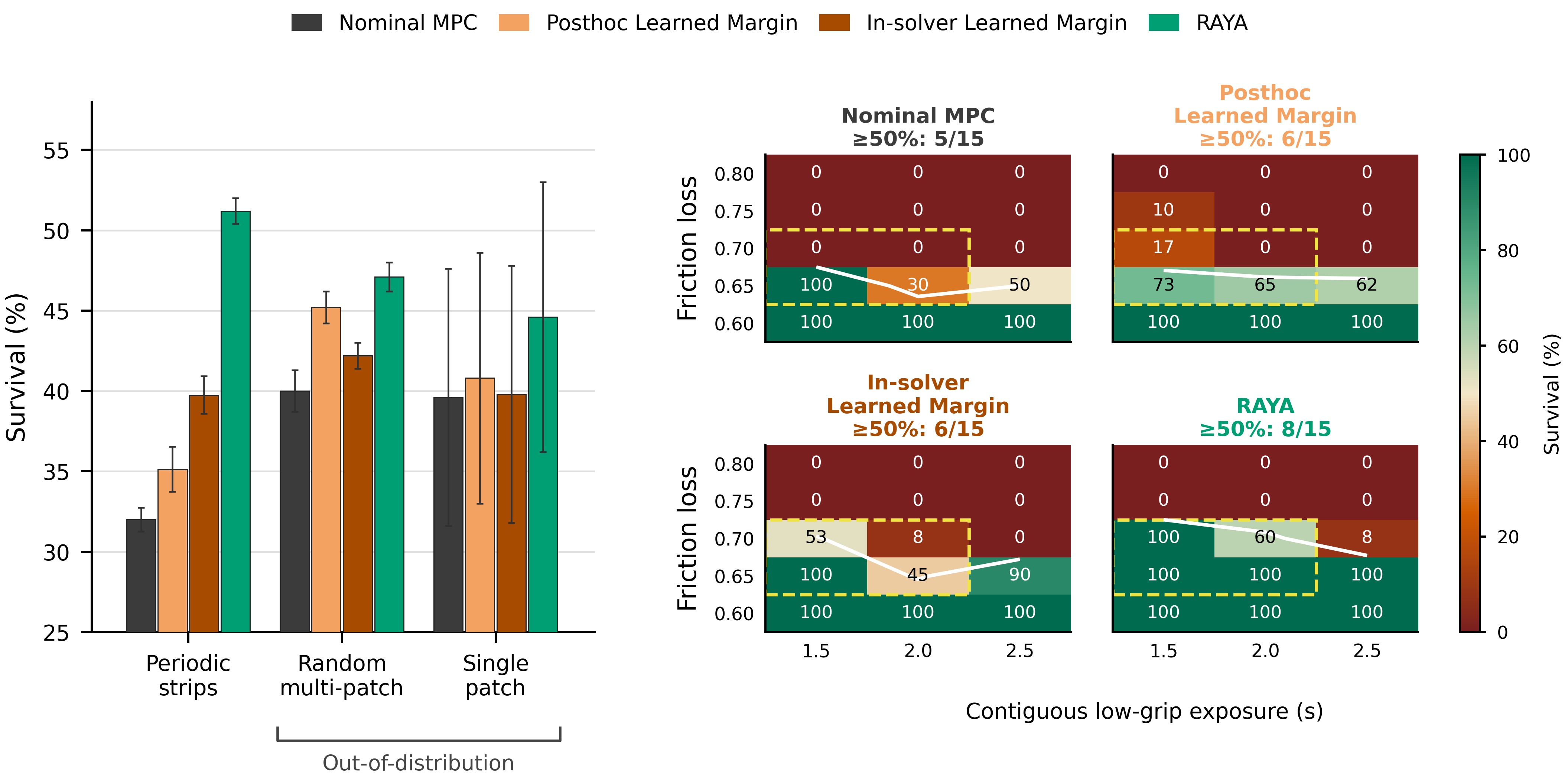}
        \vspace{-8pt}
    \caption{
        \textbf{F1TENTH recoverable operating envelope under latent friction loss.}
        \textbf{Left.} Survival on periodic strips and unseen random multi-patch and
        single-patch layouts. Whiskers show 95\% placement-clustered bootstrap intervals.
        \textbf{Right.} Survival over the periodic-strip friction-loss--exposure grid
        using 100 shared placements per condition. White curves mark the 50\%-survival
        boundary, and yellow boxes mark scenarios included during training.
    }
    \vspace{-8pt}
    \label{fig:f1_envelope}
\end{figure*}

Our experiments map directly to our three contributions. We first evaluate the hybrid recovery mechanism by isolating the roles of in-solver integration and objective reallocation (Sec.~\ref{sec:placement}). 
We then test whether \method{} expands the recoverable operating envelope beyond its training distribution (Sec.~\ref{sec:transfer}) and perform ablations to further isolate constraint depth, scheduling strategy, and component interaction (Sec.~\ref{sec:ablations}).  
Finally, we evaluate whether the same mechanism survives a performance-preserving embedded realization and physical deployment (Sec.~\ref{sec:hardware}).

\subsection{Methodology}
\label{sec:evaluation_protocol}

The quadrotor benchmark spans seven dimensionless disturbance multipliers from 6 to 12 across six evaluation conditions. On the Figure-8 disturbance profile in Fig.~\ref{fig:quad_envelope}A, these produce 95th-percentile downward disturbance accelerations of \(5.58\)--\(11.16~\mathrm{m\,s^{-2}}\). 
We report acceleration rather than airspeed because the simulator applies the disturbance directly as an acceleration without a calibrated aerodynamic mapping. 
The Figure-8 trajectory is used for training and Circle, Star, and Y-line trajectories, turbulent wind, and a plant with 20\% greater mass and additional drag are held out for zero-shot evaluation. 
We run 100 seeds at each disturbance level in each condition, giving 4,200 episodes per controller. An episode succeeds if the quadrotor completes all 281 control steps without violating the floor, attitude, or angular-rate limits.

The AV benchmark crosses five friction levels, \(\mu\in\{0.20,0.25,0.30,0.35,0.40\}\), with three periodic-strip exposure durations, two random multi-patch counts, and one isolated single-patch condition. This gives 30 condition cells. Each condition cell contains 100 pre-generated geometry placements that are reused across all controllers, yielding 1,500 periodic-strip, 1,000 random multi-patch, and 500 single-patch episodes, or 3,000 episodes per controller. 
Only the periodic-strip family is included during training, while the random multi-patch and single-patch families are held out for zero-shot evaluation. 
Failure is defined as sustained departure from the 0.8~m trajectory tube, spinout, or severe slip, consistent with F1TENTH safety-width scales~\cite{baumann2025forzaeth}.

In all cases, controllers do not receive true wind, friction, patch geometry, or future disturbance information unless explicitly identified as privileged. 
Learned models and scheduler parameters are fixed during evaluation and unless stated otherwise, all controllers are evaluated on identical disturbance seeds or geometry placements, and each ablation changes only the design choice being tested. 
In particular, post hoc versus in-solver comparisons use the same learned or HJ recoverability signal and change only where that signal enters the control pipeline. 
Baselines include nominal MPC, in-solver CBF and post hoc CBF filters~\cite{ames2016control,zeng2021safety}, reduced-order HJ supervisors~\cite{borquez2024safety}, a sampling-based policy-CBF filter~\cite{knoedler2025safety}, and post hoc and in-solver uses of the same learned recoverability margin. All nine controllers appear in Fig.~\ref{fig:quad_envelope}A. We call a condition empirically recoverable when at least 50\% of its episodes succeed.

\subsection{Simulation Experiments}

\subsubsection{In-solver Integration and Objective Reallocation}
\label{sec:placement}

We first evaluate \textit{where} the recoverability signal should enter the controller and \textit{when} selected task objectives should yield. As shown in Fig.~\ref{fig:quad_envelope}, on the
quadrotor, nominal MPC achieves 34.24\% survival, while a post hoc projection
using the learned margin achieves 34.45\% and is statistically indistinguishable
from nominal control. Using the same learned margin inside MPC increases survival to 37.48\%, an
improvement of 3.02 percentage points over post hoc use (95\% CI
$[+2.14, +3.90]$). Adding reallocation scheduling
further increases survival to 40.83\%. Thus, on the quadrotor, both in-solver
integration and objective reweighting contribute substantially to recovery.
For reference, in-solver CBF only reaches 36.74\%, below the in-solver learned margin.

A representative quadrotor trace shows the timing of the interventions.
The learned recoverability signal activates at 0.75~s, and the scheduler
output exceeds 0.5 at 0.85~s. At 0.85~s, the post hoc
projection first becomes infeasible. Nominal and post hoc control fail at
5.45~s, while \method{} completes the 14.05~s mission. The hardware results
in Sec.~\ref{sec:hardware} show the same pattern.

We also find that using the same reduced-order HJ value as a post hoc filter gives 34.24\%
survival, whereas incorporating it directly into MPC gives 36.43\%. Thus the benefit of in-solver integration is not specific to the learned margin, that is, the same signal is worth more inside the plan than after it, independent of how the signal is produced.

As shown in Fig.~\ref{fig:f1_envelope}, AV results using an F1TENTH~\cite{o2020f1tenth} car show a different balance between the two components. Nominal MPC achieves 35.93\% survival, a post hoc learned projection achieves 39.43\%, and in-solver use of the learned margin reaches 40.57\%. Adding reallocation scheduling raises survival to 48.73\%. Here, in-solver integration provides a smaller improvement, while objective reweighting accounts for the larger gain. That is, the learned margin identifies deteriorating recoverability, but effective recovery depends primarily on changing the tracking priorities before control authority is exhausted.

\subsubsection{The Recoverable Envelope Inside and Outside the Training Distribution}
\label{sec:transfer}

Fig.~\ref{fig:quad_envelope}A shows how the quadrotor benefit varies with
disturbance severity. Under mild wind, all controllers perform similarly. At
the strongest disturbances, survival collapses across controllers as the required correction
approaches the available actuation limit. The largest separation occurs
between these extremes, where recovery remains physically possible but
requires earlier intervention. An offline analysis of the
Figure-8 survival curves shows that \method shifts the interpolated 50\%-survival threshold from
\(7.335\) to \(7.950\,\mathrm{m\,s^{-2}}\) of 95th-percentile injected
downward acceleration, an 8.4\% increase, and raises the  normalized
survival-curve area from 37.17\% to 45.92\%. At disturbance
multiplier~8, survival increases from 47\% to 62\%.

We next evaluate all nine controllers zero-shot on the held-out
trajectory, wind, and plant families of Sec.~\ref{sec:evaluation_protocol}.
Only controllers improving over nominal MPC in every family are shown
in Fig.~\ref{fig:quad_envelope}B, and every controller that meets this
criterion places recoverability inside the solver.  \method shows an 18.9\% relative gain across all held-out families, and under turbulent wind survival rises
by up to 81.8\% in the regime where nominal control begins to exhaust actuation authority.

The AV experiments test the same question under unobserved
friction loss. Without retraining, \method achieves 51.2\%,
47.1\%, and 44.6\% survival on the periodic, random multi-patch, and
single-patch families, respectively, with the highest point estimate in each
family (Fig.~\ref{fig:f1_envelope}, left).

The friction-loss--exposure grid (Fig.~\ref{fig:f1_envelope}, right) shows
how \method{} expands the recoverable operating envelope. Under the 50\%-survival criterion, nominal MPC is recoverable in 5/15
conditions, while the post hoc and in-solver learned-margin controllers are
recoverable in 6/15 each. \method{} expands this to 8/15 conditions. Among the baselines, the in-solver
learned margin sustains 50\% survival to an interpolated exposure of
1.515~s. Nominal and post hoc control are already below 50\% survival at the
shortest tested exposure, so their thresholds lie below 1.5~s. \method{}
extends the 50\%-survival exposure to 2.096~s, a 38.4\% increase over the
strongest baseline.

\subsubsection{Ablation Studies}
\label{sec:ablations}

Table~\ref{tab:horizon_ablation} holds the learned models and scheduler
fixed and varies only the number \(K\) of prediction stages at which the
recoverability constraint is imposed. On the quadrotor, \(K=1\) performs
slightly better than \(K=3\) for both the learned margin-only controller and \method.
On the AV, \(K=1\) and \(K=5\) give nearly identical results. We therefore
find no benefit from repeatedly imposing the learned constraint deeper into
the MPC horizon. 

\begin{table}[t]
    \centering
    \caption{
        \textbf{Recoverability-constraint depth ablation.}
        Only \(K\) varies; learned components are fixed.
        }

    \label{tab:horizon_ablation}
    \scriptsize
    \setlength{\tabcolsep}{1.6pt}
    \begin{tabular}{@{}llcc@{}}
        \toprule
        \textbf{Platform} & \textbf{Configuration} &
        \shortstack{\textbf{Survival}\\\textbf{(95\% CI)}} &
        \shortstack{\textbf{Mean}\\\textbf{Reallocation}} \\
        \midrule
        Quadrotor & Margin, $K=1$ & 37.48 $[33.36,41.67]$ & 0.000 \\
        Quadrotor & Margin, $K=3$ & 36.33 $[32.29,40.57]$ & 0.000 \\
        Quadrotor & \textbf{\method, $K=1$} &
        \textbf{40.83} $\mathbf{[36.55,45.24]}$ & 0.698 \\
        Quadrotor & \method, $K=3$ & 39.98 $[35.74,44.36]$ & 0.702 \\
        AV & \textbf{\method, $K=1$} &
        \textbf{48.73} $\mathbf{[46.93,50.53]}$ & 0.271 \\
        AV & \method, $K=5$ & 48.70 $[46.90,50.50]$ & 0.271 \\
        \bottomrule
    \end{tabular}
\end{table}

Table~\ref{tab:scheduler_frontier} tests whether the gains come simply from
using recovery-oriented weights vs. from changing them conditionally.
As a fixed-weight baseline, we grid-search
\(\bar w\) on held-out validation rollouts and use the
best selected value for each embodiment: \(\bar w=0.75\) for the quadrotor
and \(\bar w=0.25\) for the AV. On the quadrotor, \method and always-on
scheduling achieve nearly identical survival (40.83\% and 40.88\%), but
\method uses 30.2\% less mean reallocation and reduces XY tracking error by
27.5\% on episodes completed by both controllers. On the AV, conditional
scheduling provides a larger survival benefit. \method reaches 48.73\%,
compared with 44.77\% for always-on scheduling. At the same time \method uses mean reallocation
0.271 vs. 1.0 and reduces tracking error by 22.2\%.

\begin{table}[t]
    \centering
     \caption{
      \textbf{Reallocation-scheduling ablation.}
      Only the reallocation schedule varies; the learned margin and reweighting rule are fixed.
      Static reallocation is selected on validation rollouts; lower mean reallocation indicates less
      reweighting.
      }

    \label{tab:scheduler_frontier}
    \scriptsize
    \setlength{\tabcolsep}{1.5pt}
    \begin{tabular}{@{}p{0.35\columnwidth}ccc@{}}
        \toprule
        \textbf{Quadrotor scheduler} & \textbf{Survival} &
        \shortstack{\textbf{XY error}\\\textbf{(m)}} &  \shortstack{\textbf{Mean}\\\textbf{Reallocation}} \\
        \midrule
        No reallocation (\(\bar w=0\)) & 37.48 & 0.295 & 0.000 \\
        Handcrafted ramp & 40.31 & 0.321 & 0.467 \\
        Validated static (\(\bar w=0.75\)) & 40.14 & 0.446 & 0.750 \\
        \textbf{\method} & \textbf{40.83} & \textbf{0.390} & \textbf{0.698} \\
        Always-on (\(\bar w=1\)) & \textbf{40.88} & 0.537 & 1.000 \\
        \midrule
        \textbf{AV Scheduler} & \textbf{Survival} &
        \shortstack{\textbf{Track error}\\\textbf{(m)}} & \shortstack{\textbf{Mean}\\\textbf{Reallocation}} \\
        \midrule
        No reallocation (\(\bar w=0\)) & 40.57 & 0.158 & 0.000 \\
        Handcrafted ramp & 43.43 & 0.170 & 0.068 \\
        Validated static (\(\bar w=0.25\)) & 41.77 & 0.162 & 0.250 \\
        \textbf{\method} & \textbf{48.73} & \textbf{0.161} & \textbf{0.271} \\
        Always-on (\(\bar w=1\)) & 44.77 & 0.207 & 1.000 \\
        \bottomrule
    \end{tabular}
\end{table}

We further show the two mechanisms are not merely additive but also synergistic. On the quadrotor, scheduling without the learned margin reaches 34.83\%
survival, only 0.59 percentage points above nominal MPC. The in-solver
learned margin alone reaches 37.48\%, while combining both components
reaches 40.83\%, a 2.76-percentage-point gain beyond the sum of
their individual improvements. On the AV, we compare against a privileged
controller with a 300~ms preview of the true friction, which reaches 47.50\%
survival with mean reallocation 0.503. With no friction
preview, \method reaches 48.73\% survival while using 46.1\% less mean reallocation
(0.271 vs.\ 0.503).

\begin{figure*}[!t]
    \centering
    \includegraphics[width=0.75\textwidth]{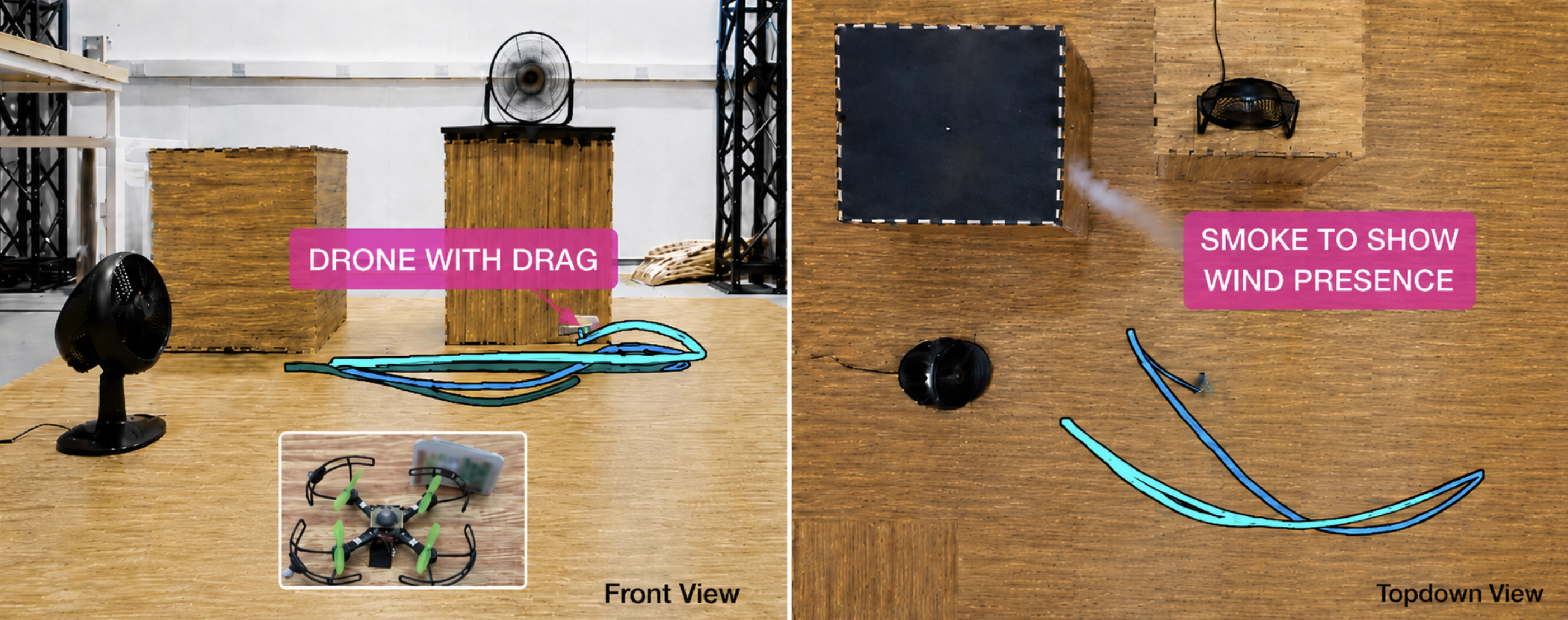}
    \vspace{-4pt}
    \caption{
        \textbf{Six-cycle onboard recovery testbed.}
        Both conditions use the same Figure-8 mission under two-fan airflow.
        The aerodynamic-mismatch condition adds a 3~g attachment, while the motor-loss
        condition applies an unmodeled 40\% reduction to the front-left motor command.
    }
    \vspace{-8pt}
    \label{fig:limo_in_action}
\end{figure*}

\subsection{Embedded Hardware Deployment}
\label{sec:hardware}

We deploy \method{} fully onboard the STM32F405 microcontroller (MCU) of a
Crazyflie 2.1 Brushless~\cite{giernacki2017crazyflie} at 20~Hz using five
ADMM iterations and a 50~ms control deadline. Each control step adds one recoverability-network evaluation and gradient,
one scheduler evaluation, one learned recoverability constraint, and updates
to selected tracking weights. Each controller is tested in five flights under each
of the two conditions, aerodynamic mismatch
and unmodeled motor loss, under the same two-fan airflow (see Fig.~\ref{fig:limo_in_action}). A mission is
successful if all six Figure-8 cycles are completed and the vehicle lands
under control. All controllers retain their nominal model and
constraints and receive no measurement or model of the disturbance or mismatch.

Table~\ref{tab:hardware_outcomes} shows a clear separation between the
controllers. \method completes all five flights in both conditions, totaling
30/30 commanded cycles per condition. Nominal MPC, in-solver CBF, and
the post hoc learned controller do not complete a single six-cycle mission in
either condition. In one motor-loss flight, \method contacts the floor after
3.286 cycles, recovers, completes the remaining trajectory, and lands under
control. Mission completion under motor loss is therefore 5/5, with 4/5
flights completed without contact.

\begin{table}[t]
    \centering
    \caption{
        \textbf{Real-world performance over five flights per condition.}
        Both conditions use the same six-cycle Figure-8 mission under two-fan airflow.
        Reported cycle counts and failure times are averaged over five flights.
    }
    \label{tab:hardware_outcomes}
    \scriptsize
    \setlength{\tabcolsep}{3.0pt}

    \begin{tabular}{@{}lccc@{}}
        \toprule
        \textbf{Method} &
        \shortstack{\textbf{Mission}\\\textbf{complete}} &
        \shortstack{\textbf{Mean cycles}\\\textbf{completed}} &
        \shortstack{\textbf{Mean time to}\\\textbf{failure (s)}} \\
        \midrule

        \multicolumn{4}{c}{\textbf{3~g attachment + wind}} \\
        \midrule
        Nominal MPC
            & \xmark~0/5
            & 0.615/6
            & 2.44 \\
        In-solver CBF
            & \xmark~0/5
            & 1.188/6
            & 4.74 \\
        Post hoc Learned Margin
            & \xmark~0/5
            & 2.135/6
            & 8.55 \\
        \textbf{\method}
            & \cmark~\textbf{5/5}
            & \textbf{6.000/6}
            & \textbf{--} \\

        \midrule
        \multicolumn{4}{c}{\textbf{40\% M1 loss + wind}} \\
        \midrule
        Nominal MPC
            & \xmark~0/5
            & 1.326/6
            & 5.32 \\
        In-solver CBF
            & \xmark~0/5
            & 3.702/6
            & 14.85 \\
        Post hoc Learned Margin
            & \xmark~0/5
            & 2.022/6
            & 8.11 \\
        \textbf{\method}
            & \cmark~\textbf{5/5}
            & \textbf{6.000/6}
            & \textbf{--} \\

        \bottomrule
    \end{tabular}
\end{table}

We also verify that the embedded implementation preserves the behavior of
the full controller. The recoverability network and its gradient are evaluated in single precision, and a static 64-entry Riccati cache replaces online recomputation of the corresponding feedback quantities. Across more than one
million replayed states, single-precision inference produces no changes in
recoverability activation. Cache quantization changes feedback gains by at
most 1.18\%, and a paired 4,200-episode simulation audit changes only 27 paired
outcomes, yielding 40.86\% survival versus 40.83\% for continuous precision. The network and cache together occupy
~161~KiB of the MCU's 1~MiB flash.

All controllers also meet the 50~ms onboard deadline. Under aerodynamic
mismatch, the worst observed control-step latency is 30.40~ms for nominal
MPC, 31.55~ms for in-solver CBF, 42.95~ms for the post hoc learned
controller, and 44.32~ms for \method. Under motor loss, the corresponding
values are 29.35, 33.79, 42.46, and 44.52~ms. The post hoc learned controller
therefore already uses 85--86\% of the available deadline, compared with
approximately 89\% for \method, yet completes none of the ten hardware
missions while \method completes all ten.

The failure modes are also consistent with the simulation results. The
post hoc projection is infeasible for 93.2\% of attempted corrections under
aerodynamic mismatch and 97.9\% under motor loss. In-solver CBF becomes
active but still fails in every flight. Together, the replay and timing audits
show that neither numerical conversion nor additional computation explains
\method's hardware advantage. By the time the post hoc filter requests a correction, there is almost never an admissible input left to deliver it. The difference is whether recoverability shapes the plan early enough, 
and whether the task is allowed to yield.

\section{Conclusion and Future Work}
\label{sec:conclusion}
We introduced \method, built on a simple premise.
Recoverability should not be checked only after the action is chosen. It should shape the plan, and what the task is allowed to sacrifice, while recovery is still possible.
Across aerial and ground embodiments, \method improves recovery over post hoc intervention, expands
the recoverable operating envelope, and transfers without retraining to
unseen tasks, disturbances, plant shifts, and friction layouts. The same
design also transfers to embedded hardware, where \method maintains recovery
under aerodynamic mismatch and unmodeled actuator degradation.

Several questions remain. The recoverability features and recovery-critical objective channels are chosen by the designer, so while the optimizer interface transfers across embodiments, some engineering remains. 
A natural step is to learn which features predict deteriorating recoverability and which objectives should yield. 
This matters most for learned nominal policies, contact-rich manipulation, and legged locomotion, where failure precursors are high-dimensional, history-dependent, and hard to enumerate in advance and we are excited to extend \method{} to these domains.

\section{Acknowledgments}
We are grateful to James Anderson, Sylvia Herbert, and William Sharpless for countless insightful discussions that led to this project. We used LLM tools, including Codex and Claude, to assist with software implementation and proofreading. All final text and references were edited and reviewed by humans.

\bibliographystyle{styles/IEEEtran}
\bibliography{styles/IEEEabrv,refs}

\end{document}